\pdfoutput=1

\documentclass[11pt]{article}

\usepackage{emnlp2021}
\usepackage{multirow}
\usepackage{xspace}
\usepackage{amssymb}
\usepackage[font=small]{subcaption}
\usepackage[font=small]{caption}
\usepackage{graphicx}

\usepackage{times}
\usepackage{latexsym}

\usepackage[T1]{fontenc}

\usepackage[utf8]{inputenc}

\usepackage{microtype}

\newcommand{\mbert}{\texttt{mBERT}\xspace}
\newcommand{\ebert}{\texttt{E-BERT}\xspace}
\newcommand{\betobert}{\texttt{BetoBERT}\xspace}
\newcommand{\arabert}{\texttt{AraBERT}\xspace}
\newcommand{\xlmr}{\texttt{XLM-R}\xspace}
\newcommand{\xlmrtw}{\texttt{XLM-Tw}\xspace}
\newcommand{\xlmrtws}{\texttt{XLM-Tw-S}\xspace}
\newcommand{\mlvat}{\texttt{mlVAT}\xspace}
\newcommand{\mlvatnosig}{\texttt{mlVAT(w/o sig)}\xspace}
\newcommand{\supex}{\texttt{Sup}\xspace}
\newcommand{\jindex}{\texttt{JI}\xspace}

\newcommand{\macrofs}{\texttt{MaF1}\xspace}
\newcommand{\microfs}{\texttt{MiF1}\xspace}
\newcommand{\kld}{\texttt{KLDivergence}\xspace}

\title{Multilingual and Multilabel Emotion Recognition using \\ Virtual Adversarial Training}

\author{Vikram Gupta \\
  ShareChat, India \\
  \texttt{vikramgupta@sharechat.co} \\}

\begin{document}
\maketitle
\begin{abstract}

Virtual Adversarial Training (VAT) has been effective in learning robust models under supervised and semi-supervised settings for both computer vision and NLP tasks. However, the efficacy of VAT for multilingual and multilabel text classification has not been explored before. In this work, we explore VAT for multilabel emotion recognition with a focus on leveraging unlabelled data from different languages to improve the model performance. We perform extensive semi-supervised experiments on SemEval2018 multilabel and multilingual emotion recognition dataset and show performance gains of \textbf{6.2\%} (Arabic), \textbf{3.8\%} (Spanish) and \textbf{1.8\%} (English) over supervised learning with same amount of labelled data (10\% of training data). We also improve the existing state-of-the-art by \textbf{7\%}, \textbf{4.5\%} and \textbf{1\%} (Jaccard Index) for Spanish, Arabic and English respectively and perform probing experiments for understanding the impact of different layers of the contextual models.

\end{abstract}

\section{Introduction}
\label{sec:intro}
Emotion recognition is an active and crucial area of research, especially for social media platforms. Understanding the emotional state of the users from textual data forms an important problem as it helps in discovering signs of fear, anxiety, bullying, hatred etc. and maintaining the emotional health of the people and platform. With the advent of deep neural networks and contextual models, text understanding has advanced dramatically by leveraging huge amount of unlabelled data freely available on web. However, even with these advancements, annotating emotion categories is expensive and time consuming as emotion categories are highly correlated and subjective in nature and can co-occur in the same text. Psychological studies suggest that emotions like "anger" and "sadness" are co-related and co-occur more frequently than "anger" and "happiness"~\cite{plutchik1980general}.
In a multilingual setup, the annotation becomes even more challenging as annotator team are expected to be familiar with different languages and culture for understanding the emotions accurately. Imbalance in availability of the data  across languages further creates problems, especially in case of resource impoverished languages. In this work, we investigate the following key points; 
\textit{a) Can unlabelled data from other languages improve recognition performance of target language and help in reducing requirement of labelled data?} 
\textit{b) Efficacy of VAT for multilingual and multilabel setup.}

To address the aforementioned questions, we focus our experiments towards~\textit{semi-supervised} learning in a~\textit{multilingual} and~\textit{multilabel} emotion classification framework. We formulate semi-supervised Virtual Adversarial Training (VAT)~\cite{miyato2018virtual} for multilabel emotion classification using contextual models and perform extensive experiments to demonstrate that unlabelled data from other languages $L_{ul} = \{L_1, L_2, \ldots, L_n\}$ improves the classification on the target language $L_{tgt}$. We obtain competitive performance by reducing the amount of annotated data demonstrating cross-language learning. To effectively leverage the multilingual content, we use multilingual contextual models for representing the text across languages. 
We also evaluate monolingual contextual models to understand the performance differences between multilingual and monolingual models and explore the advantages of \textit{domain-adaptive} and \textit{task-adaptive} pretraining of models for our task and observe substantial gains. 

We perform extensive experiments on the SemEval2018 (Affect in Tweets: Task E-c\footnote{https://competitions.codalab.org/competitions/17751}) dataset~\cite{2018semeval} which contains tweets from Twitter annotated with 11 emotion categories across three languages - English, Spanish and Arabic and demonstrate the effectiveness of semi-supervised learning across languages. To the best of our knowledge, our study is the first one to explore semi-supervised adversarial learning across different languages for multilabel classification. In summary, the main contributions of our work are the following:
\begin{itemize}
    \item We explore Virtual Adversarial Training (VAT) for semi-supervised multilabel classification on multilingual corpus.
    \item Experiments demonstrating \textbf{6.2\%}, \textbf{3.8\%} and \textbf{1.8\%} improvements (Jaccard Index) on Arabic, Spanish  and English by leveraging unlabelled data of other languages while using 10\% of labelled samples. 
    \item Improve state-of-the-art of multilabel emotion recognition by \textbf{7\%}, \textbf{4.5\%} and \textbf{1\%} (Jaccard Index) for Spanish, Arabic and English respectively.
    \item Experiments showcasing the advantages of~\textit{domain-adaptive} and ~\textit{task-adaptive} pretraining.
\end{itemize}

\section{Related Work}
Semi-supervised learning is an important paradigm for tackling the scarcity of labelled data as it marries the advantages of supervised and unsupervised learning by leveraging the information hidden in large amount of unlabelled data along with small amount of labelled data~\cite{yang2021survey},~\cite{van2020survey}. 
Early approaches used self-training for leveraging the model's own predictions on unlabelled data to obtain additional information during training ~\cite{yarowsky1995unsupervised}~\cite{mcclosky2006effective}. ~\citet{clark2018semi} proposed cross-view training (CVT) for various tasks like chunking, dependency parsing, machine translation and reported state-of-the-art results. CVT forces
the model to make consistent predictions when
using the full input or partial input. Ladder networks~\cite{laine2016temporal}, Mean Teacher networks~\cite{tarvainen2017mean} are another way for semi-supervised learning where temporal and model-weights are ensembled. Another popular direction towards semi-supervised learning is adversarial training where the data point is perturbed with random or carefully tuned perturbations to create an adversarial sample. The model is then encouraged to maintain consistent predictions for the original sample and the adversarial sample. Adversarial training was initially explored for developing secure and robust models~\cite{goodfellow2014explaining},~\cite{xiao2018characterizing},~\cite{saadatpanah2020adversarial} to prevent attacks.~\citet{miyato2016adversarial},~\citet{cheng2019robust},~\citet{zhu2019freelb} showed that adversarial training can improve both robustness and generalization for classification tasks, machine translation and GLUE benchmark respectively.~\citet{miyato2016adversarial},~\citet{sachan2019revisiting},~\citet{miyato2018virtual} then applied the adversarial training for semi-supervised image and text classification showing substantial improvements.

Emotion recognition is an important problem and has received lot of attention from the community~\cite{yadollahi2017current}, ~\cite{sailunaz2018emotion}. The taxonomies of emotions suggested by Plutchik wheel of emotions~\cite{plutchik1980general}  and~\cite{ekman1984expression} have been used by the majority of the previous work in emotion recognition. Emotion recognition can be formulated as a \textit{multiclass} problem~\cite{scherer1994evidence},~\cite{mohammad-2012-emotional} or a \textit{multilabel} problem~\cite{2018semeval},~\cite{demszky-etal-2020-goemotions}. In the \textit{multiclass} formulation, the objective is to identify the presence of one of the emotion from the taxonomy whereas in a \textit{multilabel} setting, more than one emotion can be present in the text instance. Binary relevance approach~\cite{godbole2004discriminative} is another way to break multilabel problem into multiple binary classification problems. However, this approach does not model the co-relation between emotions. Seq2Seq approaches~\cite{yang-etal-2018-sgm},~\cite{huang2021seq2emo} solve this problem by modelling the relationship between emotions by inferring emotion in an incremental manner. An interesting direction for handling data scarcity in emotion recognition is to use distant supervision by exploiting emojis~\cite{felbo-etal-2017-using}, hashtags~\cite{mohammad-2012-emotional} or pretraining emotion specific embeddings and language models~\cite{barbieri2021xlmtwitter}, ~\cite{ghosh-etal-2017-affect}.

With the emergence of contextual models like BERT~\cite{devlin2018bert}, Roberta~\cite{liu2019roberta} etc., the field of NLP and text classification has been revolutionized as these models are able to learn efficient representations from a huge corpus of unlabelled data across different languages and domains~\cite{hassan2021cross},~\cite{barbieri2021xlmtwitter}. Social media content contains linguistic errors, idiosyncratic styles, spelling mistakes, grammatical inconsistency, slangs, hashtags, emoticons etc.~\cite{barbieri2018semeval},~\cite{derczynski2013twitter} due to which off-the-shelf contextual models may not be optimum. We use ~\textit{language-adaptive},~\textit{domain-adaptive} and~\textit{task-adaptive} pretraining which has shown performance gains~\cite{peters2019tune},~\cite{gururangan2020don},~\cite{barbieri2021xlmtwitter},~\cite{howard2018universal},~\cite{lee2020biobert} for different tasks and domains.

\section{Methodology}\label{sec:task}
We consider the task of multilabel emotion classification, where given a text $t$ $\in$ $T$ and $t = \{w_1, w_2, \ldots, w_l\}$, we predict the presence of $y$ emotion categories denoted by $\{1, 2, \ldots, y\}$. $T$ represents the corpus of all the sentences across the different languages and $w_i$ represent the tokens in the sentence. We leverage contextual models as feature extractors $\phi$ : $t_i \rightarrow x_i$, where  $x_i \in \mathbb{R}^d$ and $d$ is the dimension of the text representations and train a classifier over these representations.

\subsection{Virtual Adversarial Training (VAT)}
Virtual Adversarial Training (VAT)~\cite{miyato2018virtual} is a regularization method for learning robust representations by encouraging the models to produce similar outputs for the input data points and local perturbations. VAT creates the adversary by perturbing the input in the direction which maximizes the change in the output of the model. Since VAT does not require labels it is well suited for semi-supervised applications. Consider $x \in \mathbb{R}^d$ as the $d$ dimensional representation of the text and $y$ as the ground truth. Objective function of VAT ($L_{vadv}$) is represented as,
\begin{equation}
L_{vadv}(x,\theta) := D[p(y|x,\hat{\theta}),  p(y|x + r_{vadv}, \theta)]
\end{equation}
where, 
\begin{equation}
r_{vadv} := arg\,max D[p(y|x, \hat{\theta}),  p(y|x + r, \theta)]
\label{equation_argmax}
\end{equation}
 and $||r||_2<\epsilon$ and $r_{vadv} \in \mathbb{R}^d$. $D[p,p`]$ measures the divergence between the two probability distributions and $r_{vadv}$ is the \textit{virtual adversarial perturbation} that maximizes this divergence. In order to leverage the unlabelled data, the predictions from the current estimate of the model $\hat{\theta}$ are used as the target. However, it is not possible to exactly compute $r_{vadv}$ by a closed form solution or linear approximation as gradient $g$ (Equation \ref{equation_gradient}) with respect to r is always zero at $r$ = 0.~\citet{miyato2018virtual} propose fast approximation method to formulate $r_{adv}$ as:
\begin{equation}
    r_{vadv} \approx \epsilon \frac{g}{||g||_2}, 
\end{equation}
where, 
\begin{equation}
  g = \triangledown r D[p(y|x,\hat{\theta}), p(y|x+r,\hat{\theta)}]
\label{equation_gradient}
\end{equation}
and $r=\epsilon*q$, where $q$ is a randomly sampled unit vector. With this approximation, we can use backpropagation to compute the gradients $g$ in Equation ~\ref{equation_gradient}.
The overall training objective, $L_{VAT}$ becomes:
\begin{equation}
L_{VAT} = L_{ce} + \alpha * L_{vadv}
\end{equation}
where $L_{ce}$ is the multiclass classification loss and $L_{adv}$ is the adversarial loss. $\alpha$ is the balancing hyperparameter between the two losses. 

\subsection{Multilabel Virtual Adversarial Training (\mlvat)}
We explore VAT for multilingual contextual models and multilabel classification. For computer vision tasks, perturbing the raw pixel values to generate adversarial examples is intuitive as the input space is continuous. However, contextual models use the indices of the words as input which are not present in the continuous domain and thus perturbing them is not optimal. Perturbing an index $k$ of a word $w_k$ to $k + r_{vadv}$ would not result in a word closer to $w_k$. To overcome this problem, instead of perturbing the input, we perturb the intermediate layer of the contextual models which form a continuous representation space and allows us to use VAT with contextual models. Similar strategy for text classification was also explored by~\citet{miyato2016adversarial}. For modelling multilabel classification, we measure the divergence of multilabel outputs by Mean Square Error (MSE),
\begin{equation}
L_{vadv}(x,\theta) := MSE[p(y|x,\hat{\theta}),  p(y|x + r_{vadv}, \theta)]
\end{equation}
MSE is calculated over the logits normalized by sigmoid. This is important as the outputs in case of multilabel classification are not probability distributions across classes which renders the usage of KL-Divergence incompatible for this scenario. We also experiment by treating the probability for each emotion separately but our results demonstrate the effectiveness of Mean Square Error (MSE) for our task (Table~\ref{table:loss_function}). The overall training objective, $L_{\mlvat}$ is:
\begin{equation}
L_{\mlvat} = L_{bce} + \alpha * L_{vadv}
\end{equation}where, $L_{bce}$ is the multilabel binary cross entropy loss. We represent the text instances using monolingual/multilingual contextual representations. 
\subsection{Multilingual Semi-Supervised Setup}

\textbf{\mlvat:} For each target language $L_{tgt}$, we randomly select a percentage of samples from the training set of this language and use them as labelled examples for training. We use the remaining data of the same language and the complete dataset of the other languages $L_{ul}$ as the unlabelled set. Each training batch is created by maintaining a ratio between labelled and unlabelled examples for stable training. For the labelled set, both multilabel classification loss $L_{bce}$ and adversarial loss $L_{vadv}$ is applied. For the unlabelled examples, only the adversarial loss $L_{vadv}$ is used. 

\noindent\textbf{\supex}: We also train supervised classifiers (\supex) by using the same amount of labelled data for target language $L_{tgt}$. Supervised classifiers (\supex) act as baseline and help in measuring the gains obtained by semi-supervised learning. We vary the ratio of sampled labelled examples as 10\%, 25\%, 50\% and 100\% to study the progression of our framework across different amount of labelled data of the target language.

\subsection{Multilingual Representation} 
For leveraging cross-learning between multiple languages in a semi-supervised setup, we experiment with different multilingual models. We experiment with off-the-shelf multilingual BERT, \mbert~\cite{devlin2018bert} and \xlmr~\cite{conneau2019unsupervised} models which have been trained with corpus from multiple languages. Since we are performing emotion recognition on multilingual tweets, we evaluate the \textit{domain-adaptive} multilingual model \xlmrtw~\cite{barbieri2021xlmtwitter} trained using a 198M tweet corpus across 30 languages over the \xlmr checkpoint. For exploring the effect of \textit{task-adaptive} pretraining, we evaluate ~\xlmrtws, which is finetuned for sentiment analysis over tweets which is arguably a task related to emotion recognition. 

\subsection{Monolingual Representation} 
We also experiment with monolingual models trained over the corpus from the same language for comparison with multilingual models and setting up the baselines for each language: English BERT (\ebert)~\cite{devlin2018bert} for English, \betobert~\cite{CaneteCFP2020} for Spanish and \arabert~\cite{antoun2020arabert} for Arabic. We experiment with and without finetuning the representations to evaluate the performance of these representations out-of-the box and finetuning over our task.

\begin{table}[ht]
\centering
\begin{tabular}{|l|l|c|c|c|} 
\hline
\% & Method & \jindex & \microfs & \macrofs \\
 \hline
\multirow{2}{*}{10} & \supex & 44.05 & 57.86 & 40.91\\\cline{2-5}
&\mlvat & \textbf{46.79} & \textbf{60.36} & \textbf{44.41}\\
\hline
\multirow{2}{*}{25} & \supex  & 49.69 & 62.80 & 44.19\\\cline{2-5}
& \mlvat & \textbf{51.08} & \textbf{63.96} & \textbf{47.31}\\
\hline
\multirow{2}{*}{50} & \supex  & 53.95 & 66.26 & 48.57\\\cline{2-5}
& \mlvat & \textbf{55.11} & \textbf{66.79} & \textbf{52.52}\\
\hline
\multirow{2}{*}{100} & \supex  & 55.78 & 67.41 & 50.12\\\cline{2-5}
& \mlvat & \textbf{57.31} & \textbf{68.18} & \textbf{52.15}\\
\hline
\end{tabular}
\caption{\mlvat and Supervised (\supex) experiments on \textbf{Arabic} across different ratios of labelled examples}
\label{table:finetuned_vat_sup_ar}
\end{table}

\begin{table}[ht]

\centering
\begin{tabular}{|l|l|c|c|c|} 
 \hline
\% & Method & \jindex & \microfs & \macrofs \\
\hline
\multirow{2}{*}{10} & \supex  & 54.15 & 66.33 & 48.94\\\cline{2-5}
& \mlvat & \textbf{55.15} & \textbf{67.01} & \textbf{50.57}\\
\hline
\multirow{2}{*}{25} & \supex  & 55.11 & 66.99 & 47.83\\\cline{2-5}
& \mlvat & \textbf{56.54} & \textbf{68.52} & 51.18\\
\hline
\multirow{2}{*}{50} & \supex  & 57.20 & 69.14 & \textbf{54.14}\\\cline{2-5}
& \mlvat & \textbf{58.67} & \textbf{70.03} & \textbf{51.55}\\
\hline
\multirow{2}{*}{100} & \supex  & 59.78 & 71.19 & 53.43\\\cline{2-5}
& \mlvat& \textbf{60.70} & \textbf{71.90} & \textbf{56.10}\\
\hline
\end{tabular}
\caption{\mlvat and Supervised (\supex ) experiments on \textbf{English} across different ratios of labelled examples}
\label{table:finetuned_vat_sup_en}
\end{table}

\begin{table}[ht]
\centering
\begin{tabular}{|l|l|c|c|c|} 
 \hline
\% & Method & \jindex & \microfs & \macrofs \\
 \hline
\multirow{2}{*}{10} & \supex  & 44.36 & 53.17 & 38.28\\\cline{2-5}
& \mlvat & \textbf{46.05} & \textbf{54.83} & \textbf{42.49}\\
\hline
\multirow{2}{*}{25} & \supex  & \textbf{52.89} & \textbf{61.30} & 48.99\\\cline{2-5}
& \mlvat & 52.05 & 60.17 & \textbf{49.15}\\
\hline
\multirow{2}{*}{50} & \supex  & 55.17 & 63.20 & 51.70\\\cline{2-5}
& \mlvat  & \textbf{55.70} & \textbf{63.39} & \textbf{54.19}\\
\hline
\multirow{2}{*}{100} & \supex  & \textbf{57.04} & \textbf{65.31} & 51.53\\\cline{2-5}
& \mlvat  & 56.89 & 64.89 & \textbf{51.77}\\
\hline
\end{tabular}
\caption{\mlvat and Supervised (\supex) experiments on \textbf{Spanish} across different ratios of labelled examples}
\label{table:finetuned_vat_sup_es}
\end{table}

\subsection{Dataset and Evaluation}
We evaluate on the SemEval2018 dataset (Affect in Tweets: Task E-c)~\cite{2018semeval} dataset. The dataset consists of tweets scraped from twitter in English, Spanish and Arabic. Each tweet is annotated with the presence of 11 emotions \textit{anger, anticipation, disgust, fear, joy, love, optimism, pessimism, sadness, surprise and trust}. Some tweets are neutral and do not have the presence of any emotion. The dataset has 3 splits - train, dev and test (Table~\ref{table:dataset}). Following~\citet{2018semeval}, we measure the multilabel accuracy using Jaccard Index (\jindex), Macro F1 (\macrofs) and Micro F1 (\microfs) scores~\cite{chinchor-1992-muc} over the test set of these languages.

\section{Semi-Supervised Experiments}

We select a percentage (10\%, 25\%, 50\%, 100\%) of the data from the target language as labelled data and use the remaining data from same language along with data of other languages as the unlabelled data. In Table~\ref{table:finetuned_vat_sup_ar} for Arabic, we see that by using 10\%, 25\%, 50\% and 100\% of the labelled data, \mlvat improves upon the results of training over the same amount of supervised data by \textbf{6.2\%, 2.8\%, 2.2\% and 2.7\%} (Jaccard Index;\jindex) respectively. Similar improvements are also observed on the micro F1 (\microfs) and macro F1 (\macrofs). It is interesting to note that by using only 50\% of the labelled data with unlabelled data, we are able to match the performance of supervised learning with 100\% of the data for Spanish. This shows that \mlvat is able to leverage the unlabelled data of Spanish and English for improving the performance over Arabic language. 

Similar observations on English can be made from Table~\ref{table:finetuned_vat_sup_en} also where we notice an improvement of \textbf{1.8\%, 2.6\%, 2.6\% and 2\%} on the Jaccard Index and proportional improvements on other metrices also. For English also, we note that by using 10\% of labelled data, \mlvat is able to improve on supervised results with 25\% of the data. For Spanish, \mlvat helps for the 10\% and 50\% split as reported in Table~\ref{table:finetuned_vat_sup_es} but is not able to improve all the metrics for the other splits. Overall, for majority of the languages and splits, we see that by adding unlabelled data, \mlvat improves upon the performance over supervised learning consistently and helps in decreasing the requirements for annotated data.

\begin{figure}
\centering
\begin{subfigure}[t]{0.49\columnwidth}
\centering
\includegraphics[trim=0cm 0cm 0cm 0cm, clip=true, width=\columnwidth]{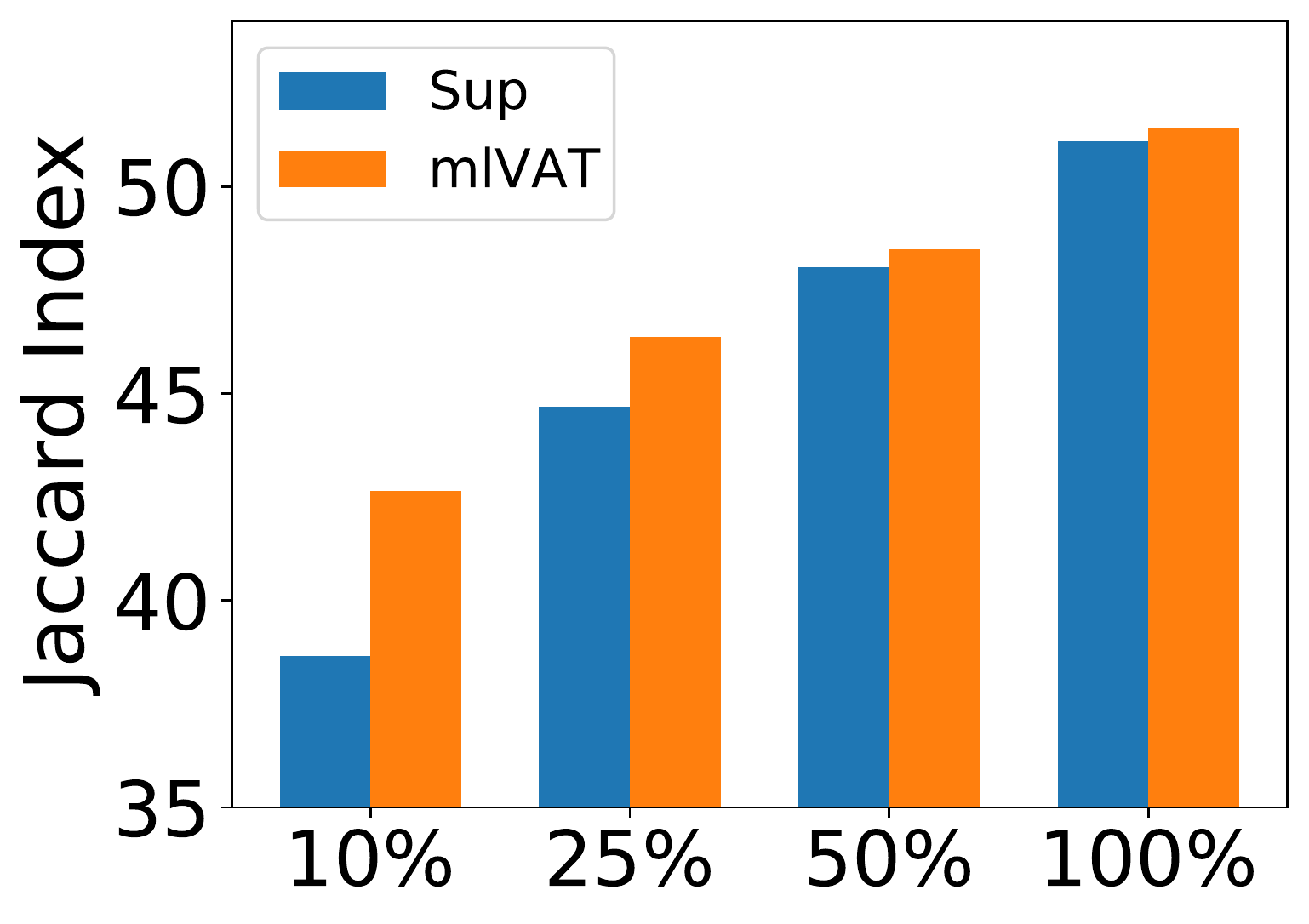}
\caption{Spanish}
\end{subfigure}
\begin{subfigure}[t]{0.49\columnwidth}
\centering
\includegraphics[trim=0cm 0cm 0cm 0cm, clip=true, width=\columnwidth]{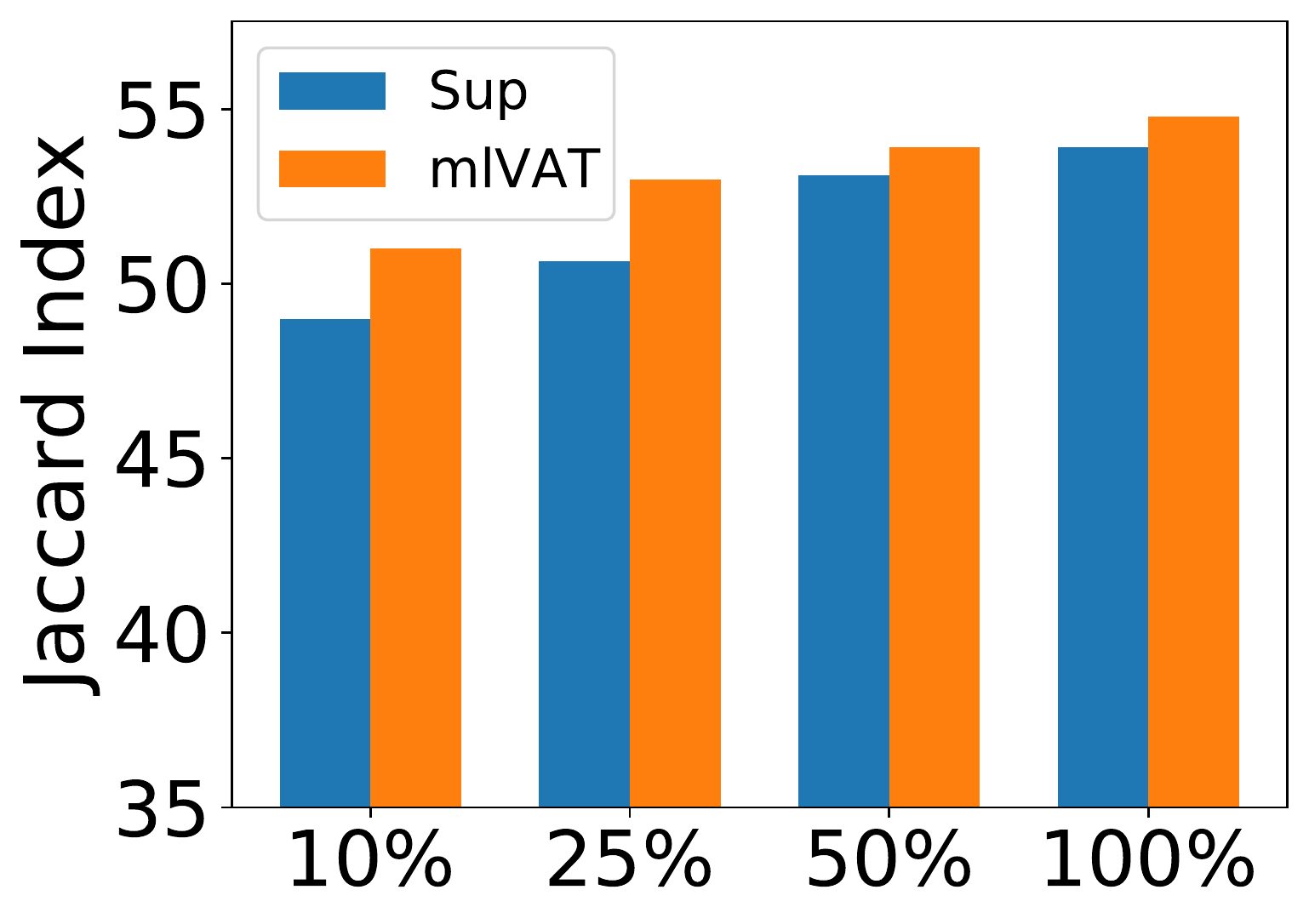}
\caption{English}
\end{subfigure}
\caption{Comparison of Jaccard Index for English and Spanish across different ratio of labelled examples with \textit{frozen backbone}. \mlvat (orange) improves upon supervised settings (\supex) (blue) for both languages.}
\label{fig:pretrained_mlvat}
\end{figure}

\noindent\textbf{Frozen backbone:}
We perform semi-supervised experiments with frozen backbone to investigate the effect of \mlvat on the backbone and classification head. We repeat similar experiments as in previous sections for Spanish and English, but freeze the backbone and only train the classification head. From the Figure~\ref{fig:pretrained_mlvat}, we can observe that \mlvat consistently improves the performance for both languages over all the splits. This demonstrates that the performance gains are backbone-agnostic allowing for application of \mlvat on other backbones also.

\begin{table}[ht]
\centering
\begin{tabular}{|l|c|c|c|} 
\hline
Loss & \jindex & \microfs & \macrofs \\
\hline
\mlvat & 55.2 & 67.1 & 50.6\\
\hline
\mlvatnosig & 50.7 & 63.5 & 41.6\\
\hline
\kld & 21.9 & 35.9 & 34.1\\
\hline
\end{tabular}
\caption{Comparison of loss functions on English with 10\% labelled data}
\label{table:loss_function}
\end{table}

\noindent\textbf{Loss Function:}
We evaluate Mean Square Loss (\mlvat), MSE without sigmoid and KL-divergence (\kld) loss in Table~\ref{table:loss_function}. MSE in presence of sigmoid shows superior performance than the other loss functions. The superior performance can be attributed to the normalization of the logits which encourages more stable activations and training. For experimenting with KL-divergence, we interpreted the normalized logits as probabilities but observed substantially poorer performance. We used English language with 10\% of labelled examples and \xlmrtw model for these experiments.

\begin{table}[ht]
\centering
\begin{tabular}{|c|c|c|c|c|c|} 
\hline
Ratio & 1 & 2 & 3 & 4 & 5 \\ 
\hline
\jindex & 55.1 & 54.4 & \textbf{55.2} & 53.6 & 52.9 \\
\hline
\microfs & 66.9 & 66.4 & \textbf{67.0} & 65.8 & 65.3 \\
\hline
\macrofs & 50.0 & \textbf{50.9} & 50.8 & 50.5 & 47.0 \\
\hline
\end{tabular}
\caption{Comparison of batch ratios on English with 10\% labelled data}
\label{table:batch_ratio}
\end{table}

\noindent\textbf{Unlabelled Batch Ratio:}
In Table~\ref{table:batch_ratio}, we study the impact of ratio of the batch size of the unlabelled examples while keeping the batch size of the labelled data fixed. At higher ratios, the adversarial loss overpowers the supervised learning resulting in a performance drop. However, for the lower ratios, the we did not observe a consistent trend. 

\begin{table}[ht]
\centering
\begin{tabular}{|c|c|c|c|c|c|} 
\hline
$\epsilon$ & 0.1 & 0.25 & 0.5 & 0.75 & 1 \\ 
\hline
\jindex & 54.9 & 54.9 & \textbf{55.2} & 54.7 & 54.6 \\
\hline
\microfs & 66.7 & 66.8 & \textbf{67.0} & 66.6 & 66.8 \\
\hline
\macrofs & 50.4 & 50.3 & \textbf{50.8} & 50.3 & 49.9 \\
\hline
\end{tabular}
\caption{Comparison of epsilon ($\epsilon$) values on English with 10\% labelled data}
\label{table:epsilon}
\end{table}

\noindent\textbf{Epsilon:}
We study the impact of epsilon ($\epsilon$) on the performance in Table~\ref{table:epsilon}. Higher values create more aggressive adversarial samples with high perturbation while lower values may create insufficient perturbation. From our empirical experiments, we note that 0.5 works better than the other values and we use this for all our semi-supervised experiments.

\section{Domain and Task Adaptive Pretraining}
In this section, we perform supervised learning experiments with frozen and finetuned representations by using the labelled data of each language for evaluating the performance of \textit{domain-adaptive}, \textit{task-adaptive}, monolingual and multilingual contextual models. In Table~\ref{table:pretrained_results_en},~\ref{table:pretrained_results_es} and~\ref{table:pretrained_results_ar}, we present the results for different monolingual and multilingual contextual models for the three languages with frozen backbones. We use English BERT (\ebert), \betobert and \arabert as monolingual models for English, Spanish and Arabic respectively. We note that for all the languages, \mbert performs substantially poorer than the monolingual contextual models of the respective languages. However, \xlmr which is another multilingual model performs competitive with the monolingual models which is not surprising as \xlmr has shown improvements over \mbert in other language tasks also~\cite{conneau2019unsupervised}. 

We further evaluate ~\textit{Domain-adaptive} (\xlmrtw) and~\textit{Task-adaptive} (\xlmrtws) versions of the \xlmr multilingual model and observe substantial improvements. \xlmrtws improves the Jaccard Index (\jindex) by \textbf{5.5\%, 6.5\%} and \textbf{8.4\%} for Arabic, English and Spanish respectively, highlighting the advantages of \textit{task-specific} pretraining for contextual models. \xlmrtw also improves upon \xlmr for all the languages reiterating the importance of pretraining the contextual models with domain specific data. 

\begin{table}[ht]
\centering
\begin{tabular}{|l|c|c|c|} 
 \hline
Model & \jindex & \microfs & \macrofs \\
 \hline
\xlmrtws & \textbf{52.0} & \textbf{64.4} & \textbf{47.9}\\
 \hline
\xlmrtw & 49.3 & 62.2 & 47.1\\
 \hline
\xlmr & 45.2 & 58.4 & 42.9\\
 \hline
\mbert & 37.5 & 51.2 & 36.2\\
 \hline
\arabert & 46.4 & 59.7 & 43.7\\
\hline
\end{tabular}
\caption{Performance of pretrained models on Arabic}
\label{table:pretrained_results_ar}
\end{table}

\begin{table}[ht]
\centering
\begin{tabular}{|l|c|c|c|} 
 \hline
Model & \jindex & \microfs & \macrofs \\
 \hline
\xlmrtws & \textbf{53.9} & \textbf{66.2} & \textbf{47.8}\\
 \hline
\xlmrtw & 50.6 & 63.5 & 45.9\\
 \hline
\xlmr & 48.6 & 61.9 & 45.7\\
 \hline
\mbert & 44.7 & 57.6 & 39.2\\
 \hline
\ebert & 48.2 & 61.4 & 42.9\\
\hline
\end{tabular}
\caption{Performance of pretrained models on English}
\label{table:pretrained_results_en}
\end{table}

\begin{table}[ht]
\centering
\begin{tabular}{|l|c|c|c|} 
 \hline
Model & \jindex & \microfs & \macrofs \\
 \hline
\xlmrtws & \textbf{51.1} & \textbf{60.0} & \textbf{48.8}\\
 \hline
\xlmrtw & 47.1 & 56.6 & 42.7\\
 \hline
\xlmr & 42.9 & 51.9 & 39.8\\
 \hline
\mbert & 37.0 & 44.8 & 31.2\\
 \hline
\betobert & 41.3 & 50.3 & 37.0\\
\hline
\end{tabular}
\caption{Performance of pretrained models on Spanish}
\label{table:pretrained_results_es}
\end{table}

We study the impact of finetuning the monolingual and best performing multilingual model on our task to compare the capabilities of multilingual models with monolingual after finetuning on the task. We notice that finetuning bridges the gap to some extent but still the domain adaptive multilingual \xlmrtw works better than the finetuned monolingual models for all the languages as shown in Table~\ref{table:finetuned_results_en}, ~\ref{table:finetuned_results_es} and ~\ref{table:finetuned_results_ar}. For English, the improvement is relatively moderate but for Spanish and Arabic, \xlmrtw demonstrates substantial gains. 

\subsection{Comparison with existing methods}

\noindent\textbf{English}:~\citet{alhuzali2021spanemo} (SpanEmo) use sentences along with emotion categories as input to the contextual model and use label correlation aware loss (LCA) to model correlation among emotions classes. LVC-Seq2Emo~\cite{huang2019seq2emo} propose a latent variable chain transformation and use it with sequence to emotion for modelling correlation between emotions. BinC~\cite{jabreel2019deep} transform the multilabel classification problem into binary classification problems and train a recurrent neural network over this transformed setting.~\cite{baziotis2018ntua} (NTUA) used a Bi-LSTM architecture with self-attention models over word2vec trained on large collection of twitter tweets and were winner of the task.~\citet{huang2021seq2emo} trained a sequence to emotion (Seq2Emo) encoder where the text is encoded using a bi-directional recurrent network and emotions are predicted by the decoder in an iterative fashion. Seq2Emo architecture allows for understanding the correlation between emotions.~\citet{yu2018improving} (DATN) use sentiments to improve emotion classification using bi-directional LSTM. ~\citet{meisheri-dey-2018-tcs} (TCS) uses SVM on manually engineered features.

\noindent\textbf{Spanish}: ~\citet{mulki-etal-2018-tw} (TW-StAR) used binary relevance transformation strategy over tweet features while~\citet{gonzalez-etal-2018-elirf} (ELiRF) explored preprocessing and adapted the tokeniser for Spanish tweets. MILAB was the wining entry in the SemEval2018 task.~\citet{hassan2021cross} (CER) finetuned the Spanish BERT representations (\betobert). 

\noindent\textbf{Arabic}: For Arabic,~\citet{samy2018context} (CA-GRU) extract contextual information from the tweets and uses them as context for emotion recognition using RNNs.~\citet{hassan2021cross} (CER) finetuned BERT representations.~\citet{alswaidan2020hybrid} (HEF) proposed hybrid neural network using different embeddings.~\citet{badaro-etal-2018-ema} (EMA) used preprocessing techniques like normalisation, stemming etc. 

Overall, our results improve upon the existing approaches on Jaccard Index(\jindex) by \textbf{7\%} for Spanish, \textbf{4.5\%} for Arabic and around \textbf{1\%} for English and setup a new state-of-the-art for all the three languages highlighting the efficacy of semi-supervised learning and \textit{domain-adaptive} multilingual models. 

\subsection{Crosslingual Experiments}
We combine data of all the three languages and train a combined model and test this model on the test set of each language. We notice that the combined model improves upon the performance of individual models for Arabic and Spanish (Table~\ref{table:finetuned_results_all}) while the performance of English is comparable.

\begin{table}[ht]
\centering
\begin{tabular}{|l|c|c|c|} 
 \hline
Model & \jindex & \microfs & \macrofs \\
\hline
\mlvat & \textbf{60.7} & \textbf{71.9} & 56.1\\
\hline
\xlmrtw & 59.8 & 71.2 & 53.4\\
\hline
\ebert & 59.1 & 70.4 & 53.3\\
\hline
\hline
SpanEmo & 60.1 & 71.3 & \textbf{57.8} \\
\hline
LVC-Seq2Emo & 59.2 & 70.9 & - \\
\hline
BinC & 59.0 & 69.2 & 56.4 \\
\hline
NTUA  & 58.8 & 70.1 & 52.8 \\
\hline
Seq2Emo & 58.7 & 70.1 & 51.9 \\
\hline
DATN & 58.3 & - & 54.4 \\
\hline
TCS & 58.2 & 69.3 & 53.0 \\
\hline
\end{tabular}
\caption{Results on English}
\label{table:finetuned_results_en}
\end{table}

\begin{table}[ht]
\centering
\begin{tabular}{|l|c|c|c|} 
 \hline
Model & \jindex & \microfs & \macrofs \\
\hline
\mlvat & 56.9 & 64.9 & 51.8 \\
\hline

\xlmrtw & \textbf{57.0} & \textbf{65.3} & 51.5\\
\hline
\betobert & 52.7 & 60.8 & 48.7\\
\hline
\hline
SpanEmo & 53.2 & 64.1 & 53.2 \\
\hline
CER & 52.4 & - & \textbf{53.7} \\
\hline
MILAB & 46.9 & 55.8 & 40.7 \\
\hline
ELiRF & 45.8 & 53.5 & 44.0 \\
\hline
TW-StAR & 43.8 & 52.0 & 39.2 \\
\hline
\end{tabular}
\caption{Results on Spanish}
\label{table:finetuned_results_es}
\end{table}

\begin{table}[ht]
\centering
\begin{tabular}{|l|c|c|c|} 
 \hline
Model & \jindex & \microfs & \macrofs \\
\hline
\mlvat & \textbf{57.3} & \textbf{68.2} & \textbf{52.2}\\
\hline
\xlmrtw & 55.8 & 67.4 & 50.1\\
 \hline
\arabert & 54.3 & 65.9 & 49.0\\
\hline
\hline
SpanEmo & 54.8 & 66.6 & 52.1 \\
\hline
CA-GRU & 53.2 & 64.8 & 49.5 \\
\hline
CER & 52.9 & - & 48.9 \\
\hline
HEF & 51.2 & 63.1 & 50.2 \\ 
\hline
EMA & 48.9 & 61.8 & 46.1 \\
\hline
\end{tabular}
\caption{Results on Arabic}
\label{table:finetuned_results_ar}
\end{table}

\begin{table}[ht]
\centering
\begin{tabular}{|l|c|c|c|} 
 \hline
Language & \jindex & \microfs & \macrofs \\
 \hline
EN & 59.4 & 70.6 & 55.7\\
 \hline
ES & 57.8 & 65.8 & 56.6\\
\hline
AR & 57.8 & 68.6 & 55.5\\
\hline
\end{tabular}
\caption{Experiments on the combination of languages}
\label{table:finetuned_results_all}
\end{table}

\begin{figure}
\centering
\includegraphics[trim=0cm 0cm 0cm 0cm, clip=true, width=\columnwidth]{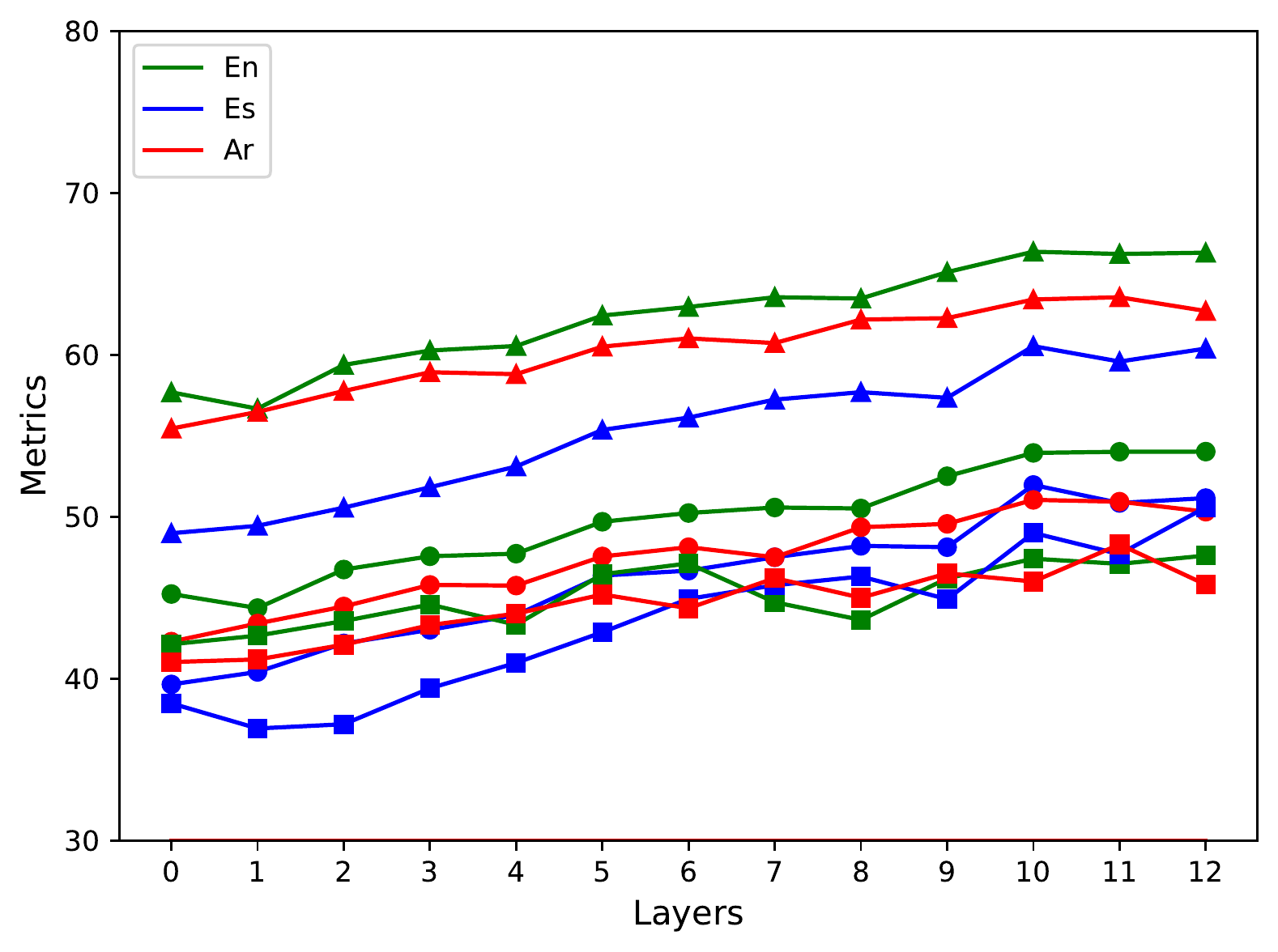}
\caption{Performance metrices across different layers for \xlmrtws. Circle, Triangle and Square represent \jindex,~\microfs and \macrofs respectively.}
\label{fig:bertology}
\end{figure}

\begin{table}[ht]
\centering
\begin{tabular}{|l|c|c|c|} 
 \hline
Train/Eval & \jindex & \microfs & \macrofs \\
\hline
Es $\rightarrow$ En & 42.9 & 54.6 & 42.6\\
Ar $\rightarrow$ En & 39.2 & 51.7 & 42.0\\
\hline
\hline
En $\rightarrow$ Ar & 49.7 & 62.3 & 45.3 \\
Es $\rightarrow$ Ar & 46.4 & 57.2 & 44.9 \\
\hline
\hline
En $\rightarrow$ Es & 44.6 & 55.9 & 41.5\\
Ar $\rightarrow$ Es & 40.0 & 51.1 & 41.9\\
\hline
\end{tabular}
\caption{Crosslingual experiments between the languages}
\label{table:crosslingual}
\end{table}
In Table~\ref{table:crosslingual}, we perform crosslingual experiments to evaluate the performance of a model trained on one language on another language. It is interesting to note that for Arabic and Spanish, the cross lingual performance is competitive with performance using some of the pretrained networks which is encouraging. We also observe that English demonstrates better crosslingual capability than Arabic and Spanish. A possible reason might be the large size of the English training dataset.

\section{Probing Experiments}
We perform experiments to evaluate the contribution of different layers of the \xlmrtws model. We extract representation of the tokens of a sentence from a particular layer of the contextual model and take an average across tokens for obtaining the representation of the sentence. We train a classifier over these sentence representations and report the results. From Figure~\ref{fig:bertology}, we note that higher layers provide better performance for all the three languages showing that the higher-order contextual information is useful for understanding the emotions in the text. Refer Appendix~\ref{Appendix:Probing} for detailed results. Similar to~\citet{tenney2019bert}, we also compute the improvement due to incrementally adding more layers to the previous layers and calculate the expected layer: 

\begin{equation}
    {E}_{\Delta}{[l]} = \frac{\sum_{l=1}^{L}l*\Delta^{(l)}}{\sum_{l=1}^{L}\Delta^{(l)}}
\label{eq:exp_layer}
\end{equation}
where, $\Delta^{(l)}$ is the change in the Jaccard Index metric when adding layer $l$ to the previous layers. We start from layer 0 and incrementally add higher layers for representing the tokens of the sentence followed by averaging for representing the whole sentence. The expected layer for English, Spanish and Arabic computes to 6.9, 6.2 and 6.8 respectively showing that higher layers are useful for the task. 
This analysis is helpful to understand the improvement achieved by adding layers to the previous layers. For all the three languages, we obtain the best results on using the average of all the layers for representing the sentences which shows that different layers encapsulate complementary information about emotions.

\begin{table}[ht]
\centering
\begin{tabular}{|l|c|c|c|} 
 \hline
 Language  & Train & Dev & Test \\ [0.5ex] 
 \hline
  English & 6838 & 886 & 3259 \\ [0.5ex] 
 \hline
 Arabic & 2278 & 585 & 1518 \\ [0.5ex] 
 \hline
 Spanish & 3561 & 679 & 2854\\ [0.5ex] 
 \hline
 \end{tabular}
\caption{SemEval2018 dataset statistics}
\label{table:dataset}
\end{table}

\section{Training Details}
We finetune the contextual models following huggingface\footnote{https://huggingface.co/} with a batch size of 8, learning rate of 2e-5 and weight decay of 0.01 using AdamW optimizer for 30 epochs. The classifier is a two layered neural network with 768 hidden dimensions and 11 output dimensions with 0.1 dropout. For \mlvat experiments, the number of examples sampled from the unlabelled set for each batch are 24, $\epsilon$ and $\alpha$ are set to 0.5 and 1 using cross validation. We apply sigmoid over the logits and train using binary cross entropy loss. We use validation set for finding optimal hyperparameters and evaluate on the test set using combination of training and validation set for training.

\section{Conclusion}
In this work, we explored semi-supervised learning using Virtual Adversarial Training (VAT) for multilabel emotion classification in a multilingual setup and showed performance improvement by leveraging unlabelled data from different languages. We used Mean Square Error (MSE) as the divergence measure for leveraging VAT for multilabel emotion classification. We also evaluated the performance of monolingual, multilingual and \textit{domain-adaptive} and \textit{task-adaptive} multilingual contextual models across three languages - English, Spanish and Arabic for multilabel and multilingual emotion recognition and obtained state-of-the-art results. We also performed probing experiments for understanding the impact of different layers of contextual models.

\section{Broader Impact and Discussion of Ethics}
In recent years, deep learning approaches have played an important role in state-of-the-art natural language processing systems. However, obtaining labelled data for training these models is expensive and time consuming, especially for multilingual and multilabel scenarios. In such case, multilingual semi-supervised and unsupervised techniques can play a pivotal role. Our work introduces a semisupervised way for detecting and understanding textual data across multiple languages. Our methods could
be used in sensitive contexts such as legal or healthcare settings, and it is essential that any work using our probe method undertake extensive quality assurance and robustness testing before using it in their setting. The datasets used in our work do not
contain any sensitive information to the best of our
knowledge.

\bibliography{anthology,custom}

\begin{thebibliography}{54}
\expandafter\ifx\csname natexlab\endcsname\relax\def\natexlab#1{#1}\fi

\bibitem[{Alhuzali and Ananiadou(2021)}]{alhuzali2021spanemo}
Hassan Alhuzali and Sophia Ananiadou. 2021.
\newblock Spanemo: Casting multi-label emotion classification as
  span-prediction.
\newblock \emph{arXiv preprint arXiv:2101.10038}.

\bibitem[{Alswaidan and Menai(2020)}]{alswaidan2020hybrid}
Nourah Alswaidan and Mohamed El~Bachir Menai. 2020.
\newblock Hybrid feature model for emotion recognition in arabic text.
\newblock \emph{IEEE Access}, 8:37843--37854.

\bibitem[{Antoun et~al.(2020)Antoun, Baly, and Hajj}]{antoun2020arabert}
Wissam Antoun, Fady Baly, and Hazem Hajj. 2020.
\newblock Arabert: Transformer-based model for arabic language understanding.
\newblock \emph{arXiv preprint arXiv:2003.00104}.

\bibitem[{Badaro et~al.(2018)Badaro, El~Jundi, Khaddaj, Maarouf, Kain, Hajj,
  and El-Hajj}]{badaro-etal-2018-ema}
Gilbert Badaro, Obeida El~Jundi, Alaa Khaddaj, Alaa Maarouf, Raslan Kain, Hazem
  Hajj, and Wassim El-Hajj. 2018.
\newblock \href {https://doi.org/10.18653/v1/S18-1036} {{EMA} at
  {S}em{E}val-2018 task 1: Emotion mining for {A}rabic}.
\newblock In \emph{Proceedings of The 12th International Workshop on Semantic
  Evaluation}, New Orleans, Louisiana. Association for Computational
  Linguistics.

\bibitem[{Barbieri et~al.(2018)Barbieri, Camacho-Collados, Ronzano, Anke,
  Ballesteros, Basile, Patti, and Saggion}]{barbieri2018semeval}
Francesco Barbieri, Jose Camacho-Collados, Francesco Ronzano, Luis~Espinosa
  Anke, Miguel Ballesteros, Valerio Basile, Viviana Patti, and Horacio Saggion.
  2018.
\newblock Semeval 2018 task 2: Multilingual emoji prediction.
\newblock In \emph{Proceedings of The 12th International Workshop on Semantic
  Evaluation}, pages 24--33.

\bibitem[{Barbieri et~al.(2021)Barbieri, Espinosa-Anke, and
  Camacho-Collados}]{barbieri2021xlmtwitter}
Francesco Barbieri, Luis Espinosa-Anke, and Jose Camacho-Collados. 2021.
\newblock {A Multilingual Language Model Toolkit for Twitter}.
\newblock In \emph{arXiv preprint arXiv:2104.12250}.

\bibitem[{Baziotis et~al.(2018)Baziotis, Athanasiou, Chronopoulou, Kolovou,
  Paraskevopoulos, Ellinas, Narayanan, and Potamianos}]{baziotis2018ntua}
Christos Baziotis, Nikos Athanasiou, Alexandra Chronopoulou, Athanasia Kolovou,
  Georgios Paraskevopoulos, Nikolaos Ellinas, Shrikanth Narayanan, and
  Alexandros Potamianos. 2018.
\newblock Ntua-slp at semeval-2018 task 1: Predicting affective content in
  tweets with deep attentive rnns and transfer learning.
\newblock \emph{arXiv preprint arXiv:1804.06658}.

\bibitem[{Cañete et~al.(2020)Cañete, Chaperon, Fuentes, Ho, Kang, and
  Pérez}]{CaneteCFP2020}
José Cañete, Gabriel Chaperon, Rodrigo Fuentes, Jou-Hui Ho, Hojin Kang, and
  Jorge Pérez. 2020.
\newblock Spanish pre-trained bert model and evaluation data.
\newblock In \emph{PML4DC at ICLR 2020}.

\bibitem[{Cheng et~al.(2019)Cheng, Jiang, and Macherey}]{cheng2019robust}
Yong Cheng, Lu~Jiang, and Wolfgang Macherey. 2019.
\newblock Robust neural machine translation with doubly adversarial inputs.
\newblock \emph{arXiv preprint arXiv:1906.02443}.

\bibitem[{Chinchor(1992)}]{chinchor-1992-muc}
Nancy Chinchor. 1992.
\newblock \href {https://aclanthology.org/M92-1002} {{MUC}-4 evaluation
  metrics}.
\newblock In \emph{{F}ourth {M}essage {U}understanding {C}onference ({MUC}-4):
  Proceedings of a Conference Held in {M}c{L}ean, {V}irginia, {J}une 16-18,
  1992}.

\bibitem[{Clark et~al.(2018)Clark, Luong, Manning, and Le}]{clark2018semi}
Kevin Clark, Minh-Thang Luong, Christopher~D Manning, and Quoc~V Le. 2018.
\newblock Semi-supervised sequence modeling with cross-view training.
\newblock \emph{arXiv preprint arXiv:1809.08370}.

\bibitem[{Conneau et~al.(2019)Conneau, Khandelwal, Goyal, Chaudhary, Wenzek,
  Guzm{\'a}n, Grave, Ott, Zettlemoyer, and Stoyanov}]{conneau2019unsupervised}
Alexis Conneau, Kartikay Khandelwal, Naman Goyal, Vishrav Chaudhary, Guillaume
  Wenzek, Francisco Guzm{\'a}n, Edouard Grave, Myle Ott, Luke Zettlemoyer, and
  Veselin Stoyanov. 2019.
\newblock Unsupervised cross-lingual representation learning at scale.
\newblock \emph{arXiv preprint arXiv:1911.02116}.

\bibitem[{Demszky et~al.(2020)Demszky, Movshovitz-Attias, Ko, Cowen, Nemade,
  and Ravi}]{demszky-etal-2020-goemotions}
Dorottya Demszky, Dana Movshovitz-Attias, Jeongwoo Ko, Alan Cowen, Gaurav
  Nemade, and Sujith Ravi. 2020.
\newblock \href {https://doi.org/10.18653/v1/2020.acl-main.372}
  {{G}o{E}motions: A dataset of fine-grained emotions}.
\newblock In \emph{Proceedings of the 58th Annual Meeting of the Association
  for Computational Linguistics}, pages 4040--4054, Online. Association for
  Computational Linguistics.

\bibitem[{Derczynski et~al.(2013)Derczynski, Ritter, Clark, and
  Bontcheva}]{derczynski2013twitter}
Leon Derczynski, Alan Ritter, Sam Clark, and Kalina Bontcheva. 2013.
\newblock Twitter part-of-speech tagging for all: Overcoming sparse and noisy
  data.
\newblock In \emph{Proceedings of the international conference recent advances
  in natural language processing ranlp 2013}, pages 198--206.

\bibitem[{Devlin et~al.(2018)Devlin, Chang, Lee, and
  Toutanova}]{devlin2018bert}
Jacob Devlin, Ming-Wei Chang, Kenton Lee, and Kristina Toutanova. 2018.
\newblock Bert: Pre-training of deep bidirectional transformers for language
  understanding.
\newblock \emph{arXiv preprint arXiv:1810.04805}.

\bibitem[{Ekman(1984)}]{ekman1984expression}
Paul Ekman. 1984.
\newblock Expression and the nature of emotion.
\newblock \emph{Approaches to emotion}, 3(19):344.

\bibitem[{Felbo et~al.(2017)Felbo, Mislove, S{\o}gaard, Rahwan, and
  Lehmann}]{felbo-etal-2017-using}
Bjarke Felbo, Alan Mislove, Anders S{\o}gaard, Iyad Rahwan, and Sune Lehmann.
  2017.
\newblock \href {https://doi.org/10.18653/v1/D17-1169} {Using millions of emoji
  occurrences to learn any-domain representations for detecting sentiment,
  emotion and sarcasm}.
\newblock In \emph{Proceedings of the 2017 Conference on Empirical Methods in
  Natural Language Processing}, pages 1615--1625, Copenhagen, Denmark.
  Association for Computational Linguistics.

\bibitem[{Ghosh et~al.(2017)Ghosh, Chollet, Laksana, Morency, and
  Scherer}]{ghosh-etal-2017-affect}
Sayan Ghosh, Mathieu Chollet, Eugene Laksana, Louis-Philippe Morency, and
  Stefan Scherer. 2017.
\newblock \href {https://aclanthology.org/P17-1059} {Affect-{LM}: A neural
  language model for customizable affective text generation}.
\newblock Vancouver, Canada. Association for Computational Linguistics.

\bibitem[{Godbole and Sarawagi(2004)}]{godbole2004discriminative}
Shantanu Godbole and Sunita Sarawagi. 2004.
\newblock Discriminative methods for multi-labeled classification.
\newblock In \emph{Pacific-Asia conference on knowledge discovery and data
  mining}, pages 22--30. Springer.

\bibitem[{Gonz{\'a}lez et~al.(2018)Gonz{\'a}lez, Hurtado, and
  Pla}]{gonzalez-etal-2018-elirf}
Jos{\'e}-{\'A}ngel Gonz{\'a}lez, Llu{\'\i}s-F. Hurtado, and Ferran Pla. 2018.
\newblock \href {https://doi.org/10.18653/v1/S18-1092} {{EL}i{RF}-{UPV} at
  {S}em{E}val-2018 tasks 1 and 3: Affect and irony detection in tweets}.
\newblock In \emph{Proceedings of The 12th International Workshop on Semantic
  Evaluation}, pages 565--569, New Orleans, Louisiana. Association for
  Computational Linguistics.

\bibitem[{Goodfellow et~al.(2014)Goodfellow, Shlens, and
  Szegedy}]{goodfellow2014explaining}
Ian~J Goodfellow, Jonathon Shlens, and Christian Szegedy. 2014.
\newblock Explaining and harnessing adversarial examples.
\newblock \emph{arXiv preprint arXiv:1412.6572}.

\bibitem[{Gururangan et~al.(2020)Gururangan, Marasovi{\'c}, Swayamdipta, Lo,
  Beltagy, Downey, and Smith}]{gururangan2020don}
Suchin Gururangan, Ana Marasovi{\'c}, Swabha Swayamdipta, Kyle Lo, Iz~Beltagy,
  Doug Downey, and Noah~A Smith. 2020.
\newblock Don't stop pretraining: adapt language models to domains and tasks.
\newblock \emph{arXiv preprint arXiv:2004.10964}.

\bibitem[{Hassan et~al.(2021)Hassan, Shaar, and Darwish}]{hassan2021cross}
Sabit Hassan, Shaden Shaar, and Kareem Darwish. 2021.
\newblock Cross-lingual emotion detection.
\newblock \emph{arXiv preprint arXiv:2106.06017}.

\bibitem[{Howard and Ruder(2018)}]{howard2018universal}
Jeremy Howard and Sebastian Ruder. 2018.
\newblock Universal language model fine-tuning for text classification.
\newblock \emph{arXiv preprint arXiv:1801.06146}.

\bibitem[{Huang et~al.(2021)Huang, Trabelsi, Qin, Farruque, Mou, and
  Zaiane}]{huang2021seq2emo}
Chenyang Huang, Amine Trabelsi, Xuebin Qin, Nawshad Farruque, Lili Mou, and
  Osmar~R Zaiane. 2021.
\newblock Seq2emo: A sequence to multi-label emotion classification model.
\newblock In \emph{Proceedings of the 2021 Conference of the North American
  Chapter of the Association for Computational Linguistics: Human Language
  Technologies}, pages 4717--4724.

\bibitem[{Huang et~al.(2019)Huang, Trabelsi, Qin, Farruque, and
  Za{\"\i}ane}]{huang2019seq2emo}
Chenyang Huang, Amine Trabelsi, Xuebin Qin, Nawshad Farruque, and Osmar~R
  Za{\"\i}ane. 2019.
\newblock Seq2emo for multi-label emotion classification based on latent
  variable chains transformation.
\newblock \emph{arXiv preprint arXiv:1911.02147}.

\bibitem[{Jabreel and Moreno(2019)}]{jabreel2019deep}
Mohammed Jabreel and Antonio Moreno. 2019.
\newblock A deep learning-based approach for multi-label emotion classification
  in tweets.
\newblock \emph{Applied Sciences}, 9(6):1123.

\bibitem[{Laine and Aila(2016)}]{laine2016temporal}
Samuli Laine and Timo Aila. 2016.
\newblock Temporal ensembling for semi-supervised learning.
\newblock \emph{arXiv preprint arXiv:1610.02242}.

\bibitem[{Lee et~al.(2020)Lee, Yoon, Kim, Kim, Kim, So, and
  Kang}]{lee2020biobert}
Jinhyuk Lee, Wonjin Yoon, Sungdong Kim, Donghyeon Kim, Sunkyu Kim, Chan~Ho So,
  and Jaewoo Kang. 2020.
\newblock Biobert: a pre-trained biomedical language representation model for
  biomedical text mining.
\newblock \emph{Bioinformatics}, 36(4):1234--1240.

\bibitem[{Liu et~al.(2019)Liu, Ott, Goyal, Du, Joshi, Chen, Levy, Lewis,
  Zettlemoyer, and Stoyanov}]{liu2019roberta}
Yinhan Liu, Myle Ott, Naman Goyal, Jingfei Du, Mandar Joshi, Danqi Chen, Omer
  Levy, Mike Lewis, Luke Zettlemoyer, and Veselin Stoyanov. 2019.
\newblock Roberta: A robustly optimized bert pretraining approach.
\newblock \emph{arXiv preprint arXiv:1907.11692}.

\bibitem[{McClosky et~al.(2006)McClosky, Charniak, and
  Johnson}]{mcclosky2006effective}
David McClosky, Eugene Charniak, and Mark Johnson. 2006.
\newblock Effective self-training for parsing.
\newblock In \emph{Proceedings of the Human Language Technology Conference of
  the NAACL, Main Conference}, pages 152--159.

\bibitem[{Meisheri and Dey(2018)}]{meisheri-dey-2018-tcs}
Hardik Meisheri and Lipika Dey. 2018.
\newblock \href {https://doi.org/10.18653/v1/S18-1043} {{TCS} research at
  {S}em{E}val-2018 task 1: Learning robust representations using
  multi-attention architecture}.
\newblock In \emph{Proceedings of The 12th International Workshop on Semantic
  Evaluation}, pages 291--299, New Orleans, Louisiana. Association for
  Computational Linguistics.

\bibitem[{Miyato et~al.(2016)Miyato, Dai, and
  Goodfellow}]{miyato2016adversarial}
Takeru Miyato, Andrew~M Dai, and Ian Goodfellow. 2016.
\newblock Adversarial training methods for semi-supervised text classification.
\newblock \emph{arXiv preprint arXiv:1605.07725}.

\bibitem[{Miyato et~al.(2018)Miyato, Maeda, Koyama, and
  Ishii}]{miyato2018virtual}
Takeru Miyato, Shin-ichi Maeda, Masanori Koyama, and Shin Ishii. 2018.
\newblock Virtual adversarial training: a regularization method for supervised
  and semi-supervised learning.
\newblock \emph{IEEE transactions on pattern analysis and machine
  intelligence}, 41(8):1979--1993.

\bibitem[{Mohammad(2012)}]{mohammad-2012-emotional}
Saif Mohammad. 2012.
\newblock \href {https://aclanthology.org/S12-1033} {{\#}emotional tweets}.
\newblock In \emph{*{SEM} 2012: The First Joint Conference on Lexical and
  Computational Semantics {--} Volume 1: Proceedings of the main conference and
  the shared task, and Volume 2: Proceedings of the Sixth International
  Workshop on Semantic Evaluation ({S}em{E}val 2012)}, pages 246--255,
  Montr{\'e}al, Canada. Association for Computational Linguistics.

\bibitem[{Mohammad et~al.(2018)Mohammad, Bravo-Marquez, Salameh, and
  Kiritchenko}]{2018semeval}
Saif Mohammad, Felipe Bravo-Marquez, Mohammad Salameh, and Svetlana
  Kiritchenko. 2018.
\newblock \href {https://aclanthology.org/S18-1001} {{S}em{E}val-2018 task 1:
  Affect in tweets}.
\newblock New Orleans, Louisiana. Association for Computational Linguistics.

\bibitem[{Mulki et~al.(2018)Mulki, Bechikh~Ali, Haddad, and
  Babao{\u{g}}lu}]{mulki-etal-2018-tw}
Hala Mulki, Chedi Bechikh~Ali, Hatem Haddad, and Ismail Babao{\u{g}}lu. 2018.
\newblock \href {https://doi.org/10.18653/v1/S18-1024} {Tw-{S}t{AR} at
  {S}em{E}val-2018 task 1: Preprocessing impact on multi-label emotion
  classification}.
\newblock In \emph{Proceedings of The 12th International Workshop on Semantic
  Evaluation}, pages 167--171, New Orleans, Louisiana. Association for
  Computational Linguistics.

\bibitem[{Peters et~al.(2019)Peters, Ruder, and Smith}]{peters2019tune}
Matthew~E Peters, Sebastian Ruder, and Noah~A Smith. 2019.
\newblock To tune or not to tune? adapting pretrained representations to
  diverse tasks.
\newblock \emph{arXiv preprint arXiv:1903.05987}.

\bibitem[{Plutchik(1980)}]{plutchik1980general}
Robert Plutchik. 1980.
\newblock A general psychoevolutionary theory of emotion.
\newblock In \emph{Theories of emotion}, pages 3--33. Elsevier.

\bibitem[{Saadatpanah et~al.(2020)Saadatpanah, Shafahi, and
  Goldstein}]{saadatpanah2020adversarial}
Parsa Saadatpanah, Ali Shafahi, and Tom Goldstein. 2020.
\newblock Adversarial attacks on copyright detection systems.
\newblock In \emph{International Conference on Machine Learning}, pages
  8307--8315. PMLR.

\bibitem[{Sachan et~al.(2019)Sachan, Zaheer, and
  Salakhutdinov}]{sachan2019revisiting}
Devendra~Singh Sachan, Manzil Zaheer, and Ruslan Salakhutdinov. 2019.
\newblock Revisiting lstm networks for semi-supervised text classification via
  mixed objective function.
\newblock In \emph{Proceedings of the AAAI Conference on Artificial
  Intelligence}, volume~33, pages 6940--6948.

\bibitem[{Sailunaz et~al.(2018)Sailunaz, Dhaliwal, Rokne, and
  Alhajj}]{sailunaz2018emotion}
Kashfia Sailunaz, Manmeet Dhaliwal, Jon Rokne, and Reda Alhajj. 2018.
\newblock Emotion detection from text and speech: a survey.
\newblock \emph{Social Network Analysis and Mining}, 8(1):1--26.

\bibitem[{Samy et~al.(2018)Samy, El-Beltagy, and Hassanien}]{samy2018context}
Ahmed~E Samy, Samhaa~R El-Beltagy, and Ehab Hassanien. 2018.
\newblock A context integrated model for multi-label emotion detection.
\newblock \emph{Procedia computer science}, 142:61--71.

\bibitem[{Scherer and Wallbott(1994)}]{scherer1994evidence}
Klaus~R Scherer and Harald~G Wallbott. 1994.
\newblock Evidence for universality and cultural variation of differential
  emotion response patterning.
\newblock \emph{Journal of personality and social psychology}, 66(2):310.

\bibitem[{Tarvainen and Valpola(2017)}]{tarvainen2017mean}
Antti Tarvainen and Harri Valpola. 2017.
\newblock Mean teachers are better role models: Weight-averaged consistency
  targets improve semi-supervised deep learning results.
\newblock \emph{arXiv preprint arXiv:1703.01780}.

\bibitem[{Tenney et~al.(2019)Tenney, Das, and Pavlick}]{tenney2019bert}
Ian Tenney, Dipanjan Das, and Ellie Pavlick. 2019.
\newblock Bert rediscovers the classical nlp pipeline.
\newblock \emph{arXiv preprint arXiv:1905.05950}.

\bibitem[{Van~Engelen and Hoos(2020)}]{van2020survey}
Jesper~E Van~Engelen and Holger~H Hoos. 2020.
\newblock A survey on semi-supervised learning.
\newblock \emph{Machine Learning}, 109(2):373--440.

\bibitem[{Xiao et~al.(2018)Xiao, Deng, Li, Yu, Liu, and
  Song}]{xiao2018characterizing}
Chaowei Xiao, Ruizhi Deng, Bo~Li, Fisher Yu, Mingyan Liu, and Dawn Song. 2018.
\newblock Characterizing adversarial examples based on spatial consistency
  information for semantic segmentation.
\newblock In \emph{Proceedings of the European Conference on Computer Vision
  (ECCV)}, pages 217--234.

\bibitem[{Yadollahi et~al.(2017)Yadollahi, Shahraki, and
  Zaiane}]{yadollahi2017current}
Ali Yadollahi, Ameneh~Gholipour Shahraki, and Osmar~R Zaiane. 2017.
\newblock Current state of text sentiment analysis from opinion to emotion
  mining.
\newblock \emph{ACM Computing Surveys (CSUR)}, 50(2):1--33.

\bibitem[{Yang et~al.(2018)Yang, Sun, Li, Ma, Wu, and
  Wang}]{yang-etal-2018-sgm}
Pengcheng Yang, Xu~Sun, Wei Li, Shuming Ma, Wei Wu, and Houfeng Wang. 2018.
\newblock \href {https://aclanthology.org/C18-1330} {{SGM}: Sequence generation
  model for multi-label classification}.
\newblock In \emph{Proceedings of the 27th International Conference on
  Computational Linguistics}. Association for Computational Linguistics.

\bibitem[{Yang et~al.(2021)Yang, Song, King, and Xu}]{yang2021survey}
Xiangli Yang, Zixing Song, Irwin King, and Zenglin Xu. 2021.
\newblock A survey on deep semi-supervised learning.
\newblock \emph{arXiv preprint arXiv:2103.00550}.

\bibitem[{Yarowsky(1995)}]{yarowsky1995unsupervised}
David Yarowsky. 1995.
\newblock Unsupervised word sense disambiguation rivaling supervised methods.
\newblock In \emph{33rd annual meeting of the association for computational
  linguistics}, pages 189--196.

\bibitem[{Yu et~al.(2018)Yu, Marujo, Jiang, Karuturi, and
  Brendel}]{yu2018improving}
Jianfei Yu, Luis Marujo, Jing Jiang, Pradeep Karuturi, and William Brendel.
  2018.
\newblock Improving multi-label emotion classification via sentiment
  classification with dual attention transfer network.
\newblock ACL.

\bibitem[{Zhu et~al.(2019)Zhu, Cheng, Gan, Sun, Goldstein, and
  Liu}]{zhu2019freelb}
Chen Zhu, Yu~Cheng, Zhe Gan, Siqi Sun, Tom Goldstein, and Jingjing Liu. 2019.
\newblock Freelb: Enhanced adversarial training for natural language
  understanding.
\newblock \emph{arXiv preprint arXiv:1909.11764}.

\end{thebibliography}
\bibliographystyle{acl_natbib}

\clearpage
\appendix

\section{Probing Experiments}
\label{Appendix:Probing}

\begin{table}[h]
\small
\begin{center}
\begin{tabular}{ |c|c|c|c| } 
 \hline
 Layers & \jindex & \microfs & \macrofs\\ [0.5ex] 
 \hline\hline
Layer 0 & 45.24 & 57.69 & 42.14 \\
Layer 1 & 44.37 & 56.68 & 42.67 \\
Layer 2 & 46.75 & 59.38 & 43.57 \\
Layer 3 & 47.57 & 60.27 & 44.59 \\
Layer 4 & 47.73 & 60.55 & 43.34 \\
Layer 5 & 49.70 & 62.43 & 46.46 \\
Layer 6 & 50.24 & 62.96 & 47.13 \\
Layer 7 & 50.58 & 63.56 & 44.73 \\
Layer 8 & 50.52 & 63.48 & 43.64 \\
Layer 9 & 52.51 & 65.11 & 46.20 \\
Layer 10 & 53.95 & 66.37 & 47.42 \\
Layer 11 & 54.02 & 66.23 & 47.10 \\
Layer 12 & 54.03 & 66.31 & 47.61 \\
\hline
\end{tabular}
\end{center}
\caption{Comparison of layer performance for English using \xlmrtws model}
\label{table:layer_wise_english}
\end{table}

\begin{table}[h]
\small
\begin{center}
\begin{tabular}{ |c|c|c|c| } 
 \hline
 Layers & \jindex & \microfs & \macrofs\\ [0.5ex] 
 \hline\hline
Layer 0 & 39.66 & 48.99 & 38.49 \\
Layer 1 & 40.43 & 49.45 & 36.94 \\
Layer 2 & 42.19 & 50.57 & 37.20 \\
Layer 3 & 43.03 & 51.83 & 39.42 \\
Layer 4 & 43.94 & 53.11 & 40.99 \\
Layer 5 & 46.38 & 55.37 & 42.88 \\
Layer 6 & 46.68 & 56.13 & 44.94 \\
Layer 7 & 47.51 & 57.24 & 45.78 \\
Layer 8 & 48.21 & 57.70 & 46.32 \\
Layer 9 & 48.13 & 57.35 & 44.92 \\
Layer 10 & 51.97 & 60.54 & 49.01 \\
Layer 11 & 50.86 & 59.59 & 47.70 \\
Layer 12 & 51.16 & 60.39 & 50.59 \\
\hline
\end{tabular}
\end{center}
\caption{Comparison of layer performance for Spanish using \xlmrtws model}
\label{table:layer_wise_spanish}
\end{table}

\begin{table}[h!]
\small
\begin{center}
\begin{tabular}{ |c|c|c|c| } 
 \hline
 Layers & \jindex & \microfs & \macrofs\\ [0.5ex] 
 \hline\hline
Layer 0 & 42.30 & 55.45 & 41.04 \\
Layer 1 & 43.42 & 56.48 & 41.20 \\
Layer 2 & 44.47 & 57.77 & 42.11 \\
Layer 3 & 45.80 & 58.93 & 43.32 \\
Layer 4 & 45.76 & 58.81 & 44.03 \\
Layer 5 & 47.56 & 60.51 & 45.21 \\
Layer 6 & 48.13 & 61.02 & 44.36 \\
Layer 7 & 47.51 & 60.73 & 46.22 \\
Layer 8 & 49.36 & 62.18 & 45.01\\
Layer 9 & 49.57 & 62.27 & 46.51 \\
Layer 10 & 51.05 & 63.42 & 46.01 \\
Layer 11 & 50.94 & 63.56 & 48.30 \\
Layer 12 & 50.32 &  62.71 & 45.82 \\
\hline
\end{tabular}
\end{center}
\caption{Comparison of layer performance for Arabic using \xlmrtws model}
\label{table:layer_wise_arabic}
\end{table}
In Table~\ref{table:layer_wise_english},~\ref{table:layer_wise_spanish} and~\ref{table:layer_wise_arabic}, we report the performance of each layer of frozen \xlmrtws model. We extract the layer representation of each token of the sentence and average them for representing the sentence. For all the languages, we note that the higher layers show superior performance.

\end{document}